\documentclass[letterpaper]{article} 
\usepackage{aaai24}  
\usepackage{times}  
\usepackage{helvet}  
\usepackage{courier}  
\usepackage[hyphens]{url}  
\usepackage{graphicx} 
\usepackage{natbib}  
\usepackage{caption} 
\usepackage{color,soul}
\usepackage{algorithm}
\usepackage{algorithmic}
\usepackage{amssymb}
\usepackage{amsmath}
\usepackage{amsfonts}
\usepackage{footmisc}
\usepackage{multirow}
\usepackage[table,xcdraw]{xcolor}
\usepackage{newfloat}
\usepackage{listings}
\DeclareCaptionStyle{ruled}{labelfont=normalfont,labelsep=colon,strut=off} 
\floatstyle{ruled}
\newfloat{listing}{tb}{lst}{}
\floatname{listing}{Listing}
\title{Emotion Rendering for Conversational Speech Synthesis with \\Heterogeneous Graph-Based Context Modeling}

\author{
    Rui Liu\textsuperscript{\rm 1}\thanks{Corrposending Author.},
    Yifan Hu\textsuperscript{\rm 1},
    Yi Ren\textsuperscript{\rm 2}, Xiang Yin\textsuperscript{\rm 2}, Haizhou Li\textsuperscript{\rm 3,4}
}
\affiliations{
    \textsuperscript{\rm 1}Inner Mongolian University, China 
    \textsuperscript{\rm 2}ByteDance\\
    \textsuperscript{\rm 3}Shenzhen Research Institute of Big Data, School of Data Science, The Chinese University of Hong Kong, Shenzhen, China\\
    \textsuperscript{\rm 4}National University of Singapore, Singapore 

    liurui\_imu@163.com, hyfwalker@163.com,  ren.yi@bytedance.com, yinxiang.stephen@bytedance.com, haizhouli@cuhk.edu.cn
}

\usepackage{bibentry}

\begin{document}

\maketitle

\begin{abstract}

Conversational Speech Synthesis (CSS) aims to accurately express an utterance with the appropriate prosody and emotional inflection within a conversational setting. While recognising the significance of CSS task, the prior studies have not thoroughly investigated the emotional expressiveness problems due to the scarcity of emotional conversational datasets and the difficulty of stateful emotion modeling. In this paper, we propose a novel emotional CSS model, termed ECSS, that includes two main components: 1) to enhance emotion understanding, we introduce a heterogeneous graph-based emotional context modeling mechanism, which takes the multi-source dialogue history as input to model the dialogue context and learn the emotion cues from the context; 2) to achieve emotion rendering, we employ a contrastive learning-based emotion renderer module to infer the accurate emotion style for the target utterance. To address the issue of data scarcity, we meticulously create emotional labels in terms of category and intensity, and annotate additional emotional information on the existing conversational dataset (DailyTalk). Both objective and subjective evaluations suggest that our model outperforms the baseline models in understanding and rendering emotions. These evaluations also underscore the importance of comprehensive emotional annotations.  Code and audio samples can be found at: {\url{https://github.com/walker-hyf/ECSS}}.

\end{abstract}
\vspace{-2mm}
\section{Introduction}


\begin{figure}[ht!]
\centering
\centerline{\includegraphics[width=1\linewidth]{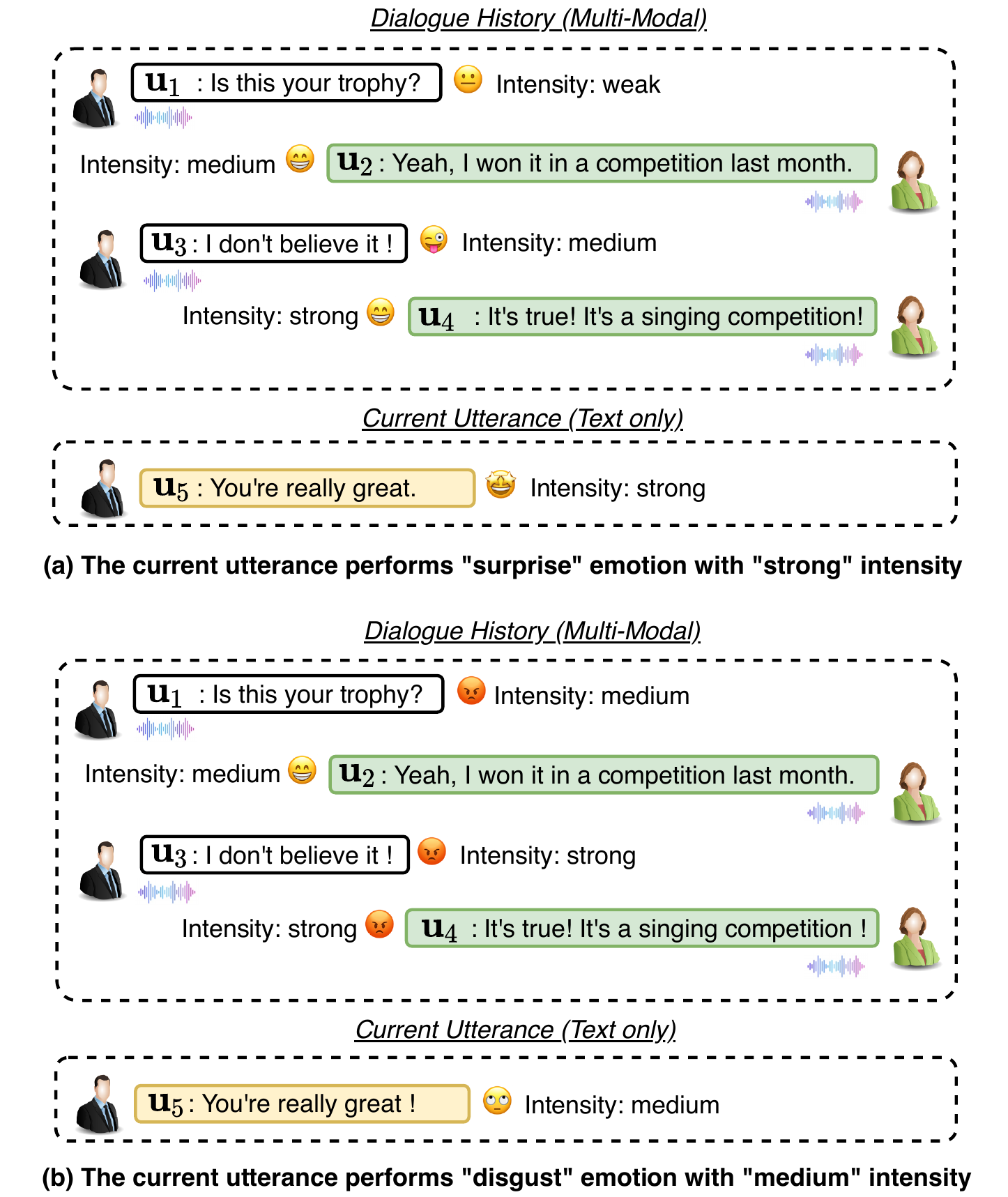}}
\vspace{-2mm}
\caption{Graphical depiction of spoken conversation, where different emotion cues in the dialogue history perform a direct impact on the emotion expression for the current utterance.
}
\vspace{-8mm}
\label{fig1}
\end{figure}

Conversational speech synthesis (CSS) aims to express a target utterance with the proper linguistic and affective prosody in a conversational context \cite{DBLP:journals/corr/abs-2005-10438}.
With the development of human-machine conversations, CSS has become an integral part of intelligent interactive systems \cite{tulshan2019survey,zhou2020design,seaborn2021voice,mctear2022conversational} and plays an important role in areas such as virtual assistants and voice agents, etc.



Unlike the speech synthesis technology for single utterance that just predict the speaking style according to its linguistic content \cite{wang2017tacotron,li2019neural,ren2021fastspeech,kim2021conditional,9271923,9420276,liu2021fasttalker,9767637} or attempt to transfer the style information from an additional reference speech \cite{wang2018style,9747987,huang2022generspeech}, CSS methods usually infer the speaking style of the target utterance according to the dialogue interaction history between two interlocutors. 
Traditional CSS works attempt to acquire the speaking style information from various aspects, such as inter and intra-speaker \cite{li2022enhancing}, and multi-modal context \cite{li2022enhancing,li2022inferring,xue2023m} dependencies modeling, etc.
For example, \citet{DBLP:journals/corr/abs-2005-10438} construct a coarse-grained context encoder at the sentence level and use the recurrent neural network (RNN) for dialogue history encoding. 
\citet{li2022enhancing} model inter-speaker and intra-speaker dependencies in a conversation by a dialog graph convolutional neural (GCN) network and summarize the output of the graph neural network using the attention mechanism.
In more recent studies, \citet{li2022inferring} propose a multi-scale relational graph convolutional network (MRGCN) to learn the dependencies in conversations at both global and local scales among the multi-modal information. 
The above researches contribute to understanding the conversational context and determining appropriate speaking styles in synthesized speeches.


However, \textbf{emotion understanding} and \textbf{emotion rendering} are largely missing in prior CSS research due to the scarcity of emotional conversational datasets and the difficulty of stateful emotion modeling.
In human-computer conversations, modeling the emotional expression in the conversational context is crucial for the speech synthesis system to generate speech with the appropriate emotional states and improve the user experience in speech-based interactions.
\textcolor{black}{As illustrated by the example in Fig.\ref{fig1}, the current utterance in Fig. \ref{fig1}(a) performs the ``Surprise'' emotion when the emotion flow of dialogue history is ``Happy $\rightarrow$ Happy $\rightarrow$ Surprise $\rightarrow$ Happy'', while in Fig. \ref{fig1}(b) performs ``Disgust'' emotion when emotion flow is ``Angry $\rightarrow$ Happy $\rightarrow$ Angry $\rightarrow$ Angry''.} We can conclude that the emotional expression in context can directly affect the speaking style for the target utterance.
In addition, humans tend to have multiple emotions with varying intensities, such as weak, medium and strong, while expressing their thoughts and feelings \cite{firdaus-etal-2020-meisd,9747098,liu2022accurate,9778970}. Therefore, emotion intensity also has an essential impact on speech expressiveness.
In a nutshell, how to fully understand the emotional cues of contextual information, and based on this, adequate emotional rendering in synthesizing conversational speech will be the focus of this paper.
Last but not least, the existing emotion-aware multimodal data \cite{busso2008iemocap,DBLP:journals/corr/abs-1810-02508,saganowski2022emognition,dias2022cross,10095836} are mostly targeted at emotion recognition scenarios, in which the speech fidelity of the audio modality is not high enough to meet the data requirements of conversational speech synthesis, resulting in the problem of data scarcity.



To address the above challenges, we propose a novel emotional CSS model, termed \textbf{ECSS}, that includes two novel mechanisms: 1) to enhance emotion understanding for the context utterances, \textit{heterogeneous graph-based emotional context modeling} module is proposed to learn the emotion cues of emotional conversational context. Specifically, given the multi-modal context in a conversation, the text, audio, speaker, emotion, and emotion intensity information are treated as the multi-source knowledge and used to build the nodes of the heterogeneous Emotional Conversational Graph called \textbf{ECG}. Then, the graph encoding module adopts the Heterogeneous Graph Transformer (HGT) \cite{hu2020heterogeneous} as the backbone to learn complex emotional dependencies in the context, and to learn the impact of such complex dependencies on the emotional expression of the current utterance;
2) to achieve the emotion rendering for the current utterance, we employ a contrastive learning-based emotion renderer module to infer the accurate emotion style for the target utterance.
Specifically, the emotion renderer takes the emotion-aware graph-enhanced node features from ECG as input and predicts the appropriate emotion, emotion intensity, and prosody features for the current utterance. This information is later aggregated with the content and speaker representations of the current utterance into the acoustic decoder to synthesize the final emotional conversational speech.
Note that the new contrastive learning losses are used to enhance the differentiation of emotion and emotional intensity expressions by drawing the same categories of emotion (or intensity) closer together and pushing different categories farther apart.
It's worth noting that, to guarantee the successful development of the ECSS model, we designed seven emotion labels (happy, sad, angry, disgust, fear, surprise, neutral), and three emotion intensity labels (weak, medium, strong) for a recent expressive conversational speech synthesis dataset, DailyTalk \cite{lee2023dailytalk}, and invited professional practitioners to annotate the labels. All annotated data will be open-sourced.
The main contributions of this paper include:
\begin{itemize}
\vspace{-1mm}
\item We propose a novel emotional conversational speech synthesis model, termed ECSS. To our best knowledge, this is the first in-depth conversational speech synthesis study that models emotional expressiveness.
\vspace{-1mm}
\item The proposed heterogeneous graph-based emotional context modeling and emotion rendering mechanisms ensure the accurate generation of emotional conversational speech in terms of emotion understanding and expression, respectively.
\vspace{-1mm}
\item Objective and subjective experiments show that the proposed model outperforms all state-of-the-art baselines in terms of emotional expressiveness.
\vspace{-1mm}
\end{itemize}


\section{Related works}



The emotion modeling for conversation has been studied in both natural language processing (NLP) and speech processing fields.
In NLP field, 
\citet{DBLP:journals/corr/abs-1909-10681} use context-aware emotion graph attention mechanisms to utilize external commonsense knowledge for emotion recognition dynamically. \citet{goel2021emotion} proposed a novel transformer encoder that adds and normalizes the word embedding with emotion embedding, thereby integrating the semantic and affective aspects of the input utterance. 
However, multi-modal information other than textual information is rarely considered in NLP.


In speech processing field, the advanced conversational speech synthesis methods use Graph Neural Networks (GNNs) models \cite{li2022enhancing,li2022inferring} to understand the multi-modal conversational context and infer the appropriate speaking style for the target utterance. For example, multi-modal features of all past utterances in the conversation, including textual information and speaking style information, are modeled by Dialogue Graph Convolutional Network (DialogueGCN) \cite{ghosal-etal-2019-dialoguegcn} to produce new representations holding richer context knowledge. However, the GNNs for CSS are designed for homogeneous graphs, in which all nodes and edges belong to the same types (such as utterance nodes from different speakers), making them infeasible to represent the natural heterogeneous structures in conversation. Especially for emotional expressions in conversation, information such as text, audio, speaker, emotion, and emotion intensity can be seen as nodes of the heterogeneous graph.

We note that there are some multi-modal conversational emotion recognition (MMCER) works that adopt heterogeneous graph networks to model the complex dependencies of contexts adequately \cite{li2022developing,song2023sunet}.
Unlike previous studies, our heterogeneous graph module has some clear differences from these works: 1) we add emotion and emotion intensity nodes into the graph structure to model the dynamic emotion cues in conversation context; 2) we adopt \textit{Heterogeneous Graph Transformer} as the backbone to encode the relations between heterogeneous nodes to learn the high-level feature representation for the constructed graph.
Note that our work is the first attempt to model the emotion understanding and rendering in conversations for CSS with heterogeneous graph networks.

\nocite{r:80, hcr:83}

\begin{figure*}[t!]
\centering
\centerline{
\includegraphics[width=1.06\linewidth]{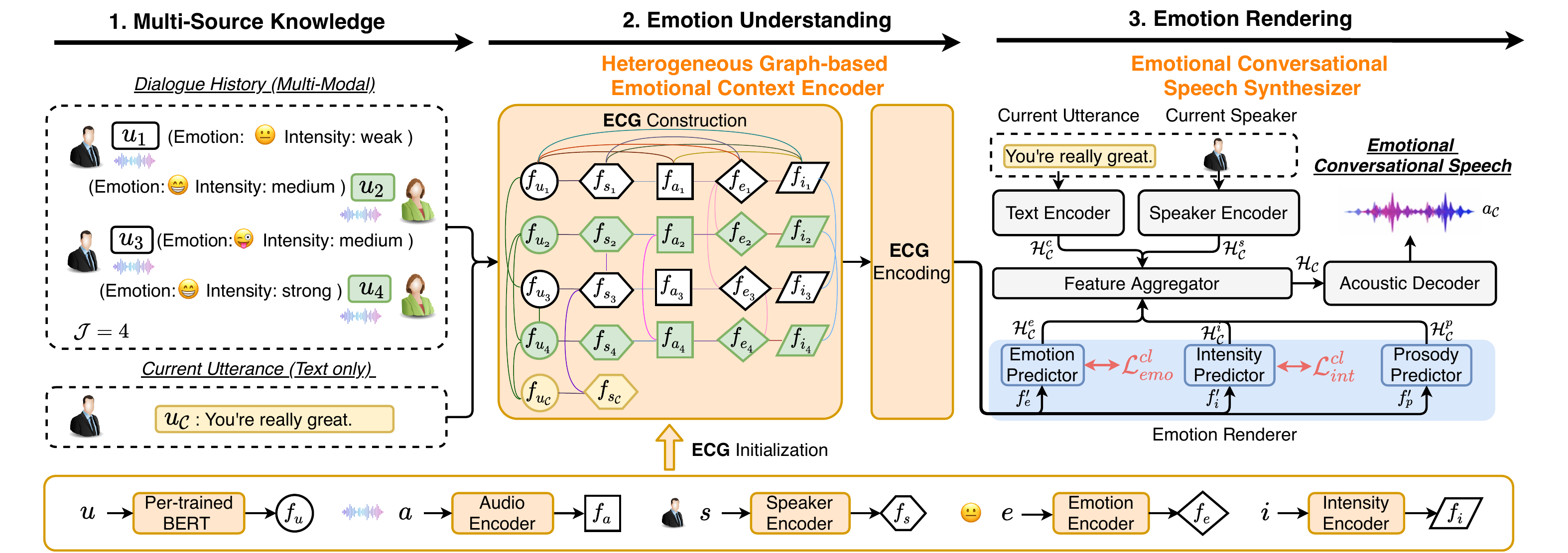}
}
\vspace{-1mm}
\caption{The overall architecture of ECSS.
1) Multi-source knowledge includes text, speaker, audio, emotion, and emotion intensity information of the multi-turn conversation; 2) Heterogeneous graph-based emotional context encoder aims to model the complex dependencies among the multi-source knowledge with the Emotional Conversational Graph (ECG), thus understanding the emotion cues in context; 3) Emotion rendering for CSS seeks to render the accurate emotion state for synthesized speech, in which \textit{Emotion Renderer} aims to infer the emotion feature according to the heterogeneous ECG encoding.
}
\vspace{-5mm}
\label{fig:model}
\end{figure*}

\section{Task Definition}
A conversation can be defined as a sequence of utterances ($utt_{1}, utt_{2}, ..., utt_{\mathcal{J}}, utt_{\mathcal{C}}$), where \{$utt_{1}, utt_{2}, ..., utt_{\mathcal{J}}$\} is the dialogue history till round $\mathcal{J}$ while $utt_{\mathcal{C}}$ means the current utterance to be synthesized. 
The task of emotional conversational speech synthesis aims to synthesize the audio $a_{\mathcal{C}}$ given the $utt_{\mathcal{C}}$ and the dialogue history \{$utt_{1}, utt_{2}, ..., utt_{\mathcal{J}}$\}. 
For the multi-modal context, each utterance $utt_{j}$ ($j \in [1, \mathcal{J}]$) in the dialogue history can be represented by five-tuples like $<$text$_{j}$, speaker$_{j}$, audio$_{j}$, emotion$_{j}$, emotion intensity$_{j}$$>$, in short for $<u_{j}$, $s_{j}$, $a_{j}$, $e_{j}$, $i_{j}$$>$. Note that the $utt_{\mathcal{C}}$ can be represented by only two-tuples like $<$text$_{\mathcal{C}}$, speaker$_{\mathcal{C}}>$, in short for $<u_{\mathcal{C}}$, $s_{\mathcal{C}}>$, since emotion and intensity information need to be generated by ECSS.
Particularly, the emotional expression of the synthesized speech $a_{\mathcal{C}}$ should confirm to the emotional conversational context characterized by the multi-modal dialogue history. To this end, the emotional CSS methods need to consider: 1) How to mine the multi-source emotional information in conversation history that is important for emotional expression; 2) How to model dynamic emotional cues in dialogue context while modeling intra-speaker, multi-modal context dependencies, etc.; 3) How to infer the appropriate emotional expression information of the current discourse based on understanding the emotional cues of the conversation.

%


\section{Methodology}
As shown in the pipeline of Fig. \ref{fig:model}, the proposed ECSS consists of three components, that are 1) \textit{Multi-source knowledge}; 2) \textit{Heterogeneous Graph-based Emotional Context Encoder} and 3) \textit{Emotional Conversational Speech Synthesizer}.
As mentioned before, the multi-modal context, including text, speaker, audio, emotion and intensity, contains natural multi-source information and can therefore be viewed as multi-source knowledge.
To enhance emotion understating, the \textit{Heterogeneous Graph-based Emotional Context Encoder} constructs a heterogeneous Emotional Conversational Graph (ECG) by considering each kind of information in the multi-source knowledge as a node, and obtains a graph-enhanced emotional contextual representation for each node after modeling the dependencies among all heterogeneous nodes. To achieve emotion rendering, the \textit{Emotional Conversational Speech Synthesizer} utilizes the graph-enhanced contextual representation in the ECG to make a reasonable prediction of the emotion expression information of the current sentence and further generates the emotional conversational speech. 
We elaborate our ECSS from the following three aspects: \textit{Heterogeneous Graph-based Emotional Context Encoder}, \textit{Emotional Conversational Speech Synthesizer} and \textit{Contrastive Learning Training Criterion}.


\subsection{Heterogeneous Graph-based Emotional Context Encoder}

As shown in the middle panel of Fig. \ref{fig:model}, the heterogeneous graph-based emotional context encoder
consists of three parts: 1) ECG Construction, constructing the heterogeneous graph with multi-source context; 2) ECG Initialization, initializing different heterogeneous nodes via their own feature representations; 3) ECG Encoding, perceiving emotion cues and generating the emotion-aware feature representation for the heterogeneous nodes. 


\subsubsection{ECG Construction} 
Unlike previous GNNs based CSS methods, we aim to introduce the multi-source knowledge, which are 5 kinds of nodes including text $f_u$, audio $f_a$, speaker $f_s$, emotion $f_e$, and intensity $f_i$ and build an emotional conversational graph or ECG $\mathcal{G = (N, E)}$, where $\mathcal{N}$ denotes the set of nodes, and $\mathcal{E}$ denotes the set of edges representing the relations between two nodes. Note that the speaker, audio and text nodes seek to introduce the basic dialogue attributes, while emotion and intensity nodes can introduce the dynamic emotion traits and bridge the emotion interaction between remote utterances.
As shown in the middle part of Fig. \ref{fig:model}, different shapes of diagrams mean different kinds of nodes. 


Considering the multi-source knowledge, We created 14 different types of edges, as shown by the different colored connecting lines shown in the middle part of Fig. \ref{fig:model}. However, due to space limits, not all the edges of the nodes are depicted. 
In a nutshell, these 14 edges connect  1) the text and each of the other nodes, 2) the audio and speaker nodes, 3) the emotion and speaker, emotion intensity, and audio nodes, and 4) emotion intensity and speaker, audio nodes. Note that all edges include past-to-future and future-to-past connections to model the bidirectional relation.

\subsubsection{ECG Initialization} 
To achieve meaningful heterogeneous graph encoding, we need to initialize all nodes with their feature representations. As shown in the middle and bottom panels of Fig. \ref{fig:model}, to take the multi-turn dialogue as an example, we employ various encoders to obtain $f_{u_j}$, $f_{s_j}$, $f_{a_j}$, $f_{e_j}$, $f_{i_j}$ ($j \in [1,4]]$) for text, speaker, audio, emotion, and intensity nodes. 

\begin{itemize}
    \item \textbf{Text Nodes}. We adopt a pre-trained BERT\footnote{\label{bert}https://huggingface.co/sentence-transformers/distiluse-base-multilingual-cased-v1} model to extract the linguistic feature: $f_{u_{j}} = {\rm BERT}(u_{j})$.

    \item \textbf{Audio Nodes}. We employ the global style token (GST) \cite{wang2018style} module, which includes a reference encoder and style token layer, as the audio encoder to extract the acoustic features contained in each audio $a_j$: $f_{a_{j}} = {\rm GST}(a_j)$.

    \item \textbf{Speaker}, \textbf{Emotion}, and \textbf{Emotion Intensity Nodes}. The speaker, emotion and intensity encoders are used to define three randomly initialized trainable parameter matrices 
    $f_{s_{j}}$, $f_{e_{j}}$, and $f_{i_j}$
    to learn two speaker identity features, seven emotion label (happiness, sadness, anger, disgust, fear, surprise, neutral) features, and three emotion intensity label (weak, medium, strong) features respectively. 
 \vspace{-2mm}
\end{itemize}

Note that the text and speaker nodes $f_{u_\mathcal{C}}$, $f_{i_\mathcal{C}}$ of current utterance are initialized in the same way as the nodes $f_{u_j}$ and $f_{s_j}$ in the dialogue history.
 
\subsubsection{ECG Encoding}
After initializing the constructed Graph, the heterogeneous graph encoding module is used to encode the emotion cues in dialogue context to obtain the graph-enhanced representations for each node.
Inspired by Heterogeneous Graph Transformer (HGT) \cite{hu2020heterogeneous}, we also adopt a three-stage emotional HGT network, that includes \textit{Heterogeneous Mutual Attention} (HMA), \textit{Heterogeneous Message Passing} (HMP), and \textit{Emotional knowledge Aggregation} (EKA) operations, to model the dependencies in emotional conversations.
Assuming that one node in the heterogeneous ECG is the target node, then any node of any type can be viewed as the source node. The goal of HGT is to aggregate information from source nodes to get a contextualized representation for the target node. 

Firstly, given a target node $\mathcal{N}_{\hat {tgt}}$ and all its neighbor source nodes $\mathcal{N}_{\hat {src}}$, the HMA mechanism maps $\mathcal{N}_{\hat {tgt}}$ into a Query vector, and $\mathcal{N}_{\hat {src}}$ into a Key vector, and calculate their dot product as attention, that indicates the importance of each source node for the target node. Then we concatenate $h$ attention heads together to get the attention vector for each node pair $\mathcal{E}_{\hat {src} \rightarrow \hat {tgt}}= \{\mathcal{N}_{\hat {src}}, \mathcal{N}_{\hat {tgt}}\}$ ($\mathcal{E}_{\hat {src} \rightarrow \hat {tgt}}$ means the edge between $\mathcal{N}_{\hat {src}}$ and $\mathcal{N}_{\hat {tgt}}$). For each target node $\mathcal{N}_{\hat {tgt}}$, we gather all attention vectors from its neighbors $\mathcal{N}_{\hat {src}}$ and conduct softmax to get the final attention score.

Secondly, HMP is computed parallel to pass the dependency information from $\mathcal{N}_{\hat {src}}$ to $\mathcal{N}_{\hat {tgt}}$. Specifically, for a pair of nodes $\mathcal{E}_{\hat {src} \rightarrow \hat {tgt}}= \{\mathcal{N}_{\hat {src}}, \mathcal{N}_{\hat {tgt}}\}$, HMP uses a linear projection to project the feature vector of $\mathcal{N}_{\hat {src}}$ into a message vector, that then followed by a matrix $W_{\hat {src} \rightarrow \hat {tgt}}$ for incorporating the edge dependency. The final step is to concatenate all $h$ message heads to get the final message vector for each node pair.

At last, the EKA module aims to aggregate the computed attention score and message vector.
We use the attention score as the weight to average the corresponding messages from all neighbor source nodes $\mathcal{N}_{\hat {src}}$ and get the emotion-augmented vector $f_{\mathcal{N}_{\hat {tgt}}}$ for the target node $\mathcal{N}_{\hat {tgt}}$. 
It's worth noting that all ECG nodes, including emotion and intensity nodes, can be treated as the source or target nodes to learn its high-level contextual information. Therefore, the EKA operation ultimately incorporates the emotional information from the dialog history into all ECG nodes. The graph-enhanced feature representation of each node incorporates the emotional cues in context.


In this way, the ECG encoding module can obtain the final emotion-aware graph-enhanced feature representation $f_{u}^{'}$, $f_{s}^{'}$, $f_{a}^{'}$, $f_{e}^{'}$ and $f_{i}^{'}$ for all text, speaker, audio, emotion and intensity nodes respectively, which can be fed into subsequent models to provide emotional information.

\subsection{Emotional Conversational Speech Synthesizer}
As shown in the right panel of Fig. \ref{fig:model}, the emotional conversational speech synthesizer consists of the following four components: \textit{Text encoder, Speaker Encoder, Emotion Renderer and Acoustic Decoder}.
The text and speaker encoders seek to encode the content and speaker identity features $\mathcal{H}^c_\mathcal{C}$ and $\mathcal{H}^s_\mathcal{C}$ for the current utterance and speaker. 
Note that the emotion renderer attempts to predict the current utterance's emotion, intensity, and prosody features $\mathcal{H}^e_\mathcal{C}$, $\mathcal{H}^i_\mathcal{C}$, and $\mathcal{H}^p_\mathcal{C}$ using the graph-enhanced node features. 
To obtain a robust feature representation  of the current utterance, the feature aggregator module set a set of trainable weight parameters
for the above five features and then output the final mixup feature $\mathcal{H}_\mathcal{C}$.
For the acoustic decoder, we use \textit{FastSpeech2} \cite{ren2021fastspeech} as the backbone, which includes the variance adaptor, mel decoder and vocoder. The variance adapter takes the $\mathcal{H}_\mathcal{C}$ as inputs to predict the duration, energy, and pitch. The Mel Decoder aims to predict the mel-spectrum features. 
Finally, a well-trained HiFi-GAN \cite{kong2020hifi} is used as the vocoder to generate speech waveform $a_\mathcal{C}$ with desired emotion style.

Note that to achieve emotion rendering, our emotion renderer extracts the encoded node features from ECG and predicts the emotion and emotion intensity feature representations of the current utterance while performing the prosody prediction.  More importantly, to achieve accurate emotion category and emotion intensity feature representation prediction, we propose contrastive learning-based emotion and emotion intensity loss functions $\mathcal{L}^{cl}$.



\vspace{-1mm}
\subsubsection{Emotion Renderer}
As shown in the blue part of Fig. \ref{fig:model}, the emotion renderer consists of emotion, intensity and prosody predictors. The prosody predictor adopts a multi-head attention layer to infer the speaking prosody information of the current utterance from the feature representations of text nodes in the dialogue history. The features of audio nodes are not used because we believe that audio and text modalities have already interacted during the ECG encoding, thus the text node already contains audio information. We use the MSE loss for the prosody predictor to constrain its training, where the target is obtained from a GST-based prosody extractor. Next, we will introduce emotion and intensity predictors.

\begin{itemize}
   \vspace{-1mm}
    \item \textbf{Emotion Predictor} uses the encoded features of the emotion nodes in dialog history to infer the emotion representation $\mathcal{H}^e_\mathcal{C}$ of the current sentence.
    It includes two convolutional layers, a bidirectional LSTM layer, and two fully connected layers. 
    \vspace{-1mm}
    \begin{equation}
        \mathcal{H}^e_\mathcal{C}= {\rm FC(BiLSTM(CNN(}f_e^{'})))
    \end{equation}
    where $f_e^{'}$ represents all emotion-type nodes in the dialogue history after ECG encoding.

 \vspace{-0.3mm}
    \item \textbf{Intensity Predictor} uses the encoded features of the intensity nodes in dialog history to infer the emotion intensity representation $\mathcal{H}^i_\mathcal{C}$ of the current sentence. It consists of two convolutional layers, a bidirectional LSTM layer, two fully connected layers, and a mean pooling layer. 
    \begin{equation}
    \begin{split}
        \mathcal{H}^i_\mathcal{C}= {\rm AvgPooling(FC_2(BiLSTM(CNN_2(}f_i^{'}))))
    \end{split}
    \end{equation}
    where $f_i^{'}$ is a universal representation of all emotion intensity nodes in the dialog history after ECG encoding.
 
\end{itemize}

\subsubsection{Contrastive Learning Training Criterion}

For emotion and intensity predictors of the emotion renderer, inspired by \cite{DBLP:journals/corr/abs-2004-11362}, we design the emotion-supervised contrastive learning losses
$\mathcal{L}^{cl}$ to motivate the emotion renderer to better distinguish different emotions categories and intensity degrees. 
Specifically, contrastive learning loss $\mathcal{L}^{cl}_{emo}$ for emotion category and  $\mathcal{L}^{cl}_{int}$ for emotion intensity share the same spirits, that is treating all examples with the same emotion category or intensity label in the batch as positive examples while different labels as negative.

For the emotion feature $\mathcal{H}^{e}_\mathcal{C}$, a batch of $K$ emotion representations is denoted as $\mathcal{H}^{K} = [\mathcal{H}^{e}_{\mathcal{C}1}, \mathcal{H}^{e}_{\mathcal{C}2}, ..., \mathcal{H}^{e}_{\mathcal{C}K}]$,
 $\mathcal{L}^{cl}_{emo}$ for $\mathcal{H}^{e}_{\mathcal{C}k}$ as follows,
\vspace{-1.5mm}
\begin{equation}
    \mathcal{L}^{cl}_{emo}= log \frac{-1}{|\mathcal{P}(k)|} \frac{\underset{\mathcal{H}^{e}_{\mathcal{C}q}\in \mathcal{P}(k)}{\sum } exp(sim(\mathcal{H}^{e}_{\mathcal{C}k},\mathcal{H}^{e}_{\mathcal{C}q} )/\tau)}{\underset{\mathcal{H}^{e}_{\mathcal{C}d}\in B(k)}{\sum} exp(sim(\mathcal{H}^{e}_{\mathcal{C}k},\mathcal{H}^{e}_{\mathcal{C}d})/\tau)}
\end{equation}
where $sim(\cdot,\cdot)$ is a cosine similarity function. $\tau$ is a scalar temperature parameter. $B(k) \equiv \mathcal{H}^{K} \backslash \{\mathcal{H}^{e}_{\mathcal{C}k} \}$ contains all representations in $\mathcal{H}^{K}$ except $\mathcal{H}^{e}_{\mathcal{C}k}$. $\mathcal{P}(k)$ is the set of positive samples that have the same emotion label with $\mathcal{H}^{e}_{\mathcal{C}k}$ in a batch.
Similarity,  $\mathcal{L}^{cl}_{int}$ for $\mathcal{H}^{i}_{\mathcal{C}k}$ as follows,

\vspace{-1.5mm}
\begin{equation}
    \mathcal{L}^{cl}_{int}= log \frac{-1}{|\mathcal{P}(k)|} \frac{\underset{\mathcal{H}^{i}_{\mathcal{C}q}\in \mathcal{P}(k)}{\sum } exp(sim(\mathcal{H}^{i}_{\mathcal{C}k},\mathcal{H}^{i}_{\mathcal{C}q} )/\tau)}{\underset{\mathcal{H}^{i}_{\mathcal{C}d}\in B(k)}{\sum} exp(sim(\mathcal{H}^{i}_{\mathcal{C}k},\mathcal{H}^{i}_{\mathcal{C}d})/\tau)}
\end{equation}
\vspace{-1.5mm}

At last, the total loss function $\mathcal{L}$ is: $
\mathcal{L} = \mathcal{L}^{cl}_{emo} + \mathcal{L}^{cl}_{int} + \mathcal{L}^{mse}_{pro} + \mathcal{L}_{fs2}$, where $\mathcal{L}_{fs2}$ refers to the acoustic feature loss of traditional FastSpeech2, including pitch, energy, and duration, as well as the mel spectrum, $\mathcal{L}^{mse}_{pro}$ indicates the MSE loss for prosody predictor. 

\begin{table*}[t!]
\centering
\caption{\label{tab:methods}\textcolor{black}{Subjective (with 95\% confidence interval) and objective results with different CSS systems. (* means the metric value achieved the suboptimal result among all systems.)}} 
\vspace{-3mm}
\resizebox{0.97\linewidth}{!}{
\begin{tabular}{lcccccc}
\hline
\textbf{Systems}       & \textbf{N-DMOS} ($\uparrow$)   & \textbf{E-DMOS} ($\uparrow$)                               & \multicolumn{1}{l}{\textbf{MAE-M} ($\downarrow$) }    & \multicolumn{1}{l}{\textbf{MAE-P ($\downarrow$)}}    & \multicolumn{1}{l}{\textbf{MAE-E ($\downarrow$)}}     & \multicolumn{1}{l}{\textbf{MAE-D ($\downarrow$)}}     \\ \hline
 No emotional context modeling  & 3.232 $\pm$ 0.030                                  & 3.100 $\pm$ 0.026                                  & 0.681                                 & 0.506                                 & 0.346                                  & 0.300                                  \\
 GRU-based                      & 3.314 $\pm$ 0.022                                 & 3.288 $\pm$ 0.011                                 & 0.675                                 & 0.506                                 & 0.352                                  & 0.296                                  \\
 Homogeneous Graph-based      & 3.384 $\pm$ 0.027                                 & 3.493 $\pm$ 0.031                                 & {\color[HTML]{393939} 0.662}          & {\color[HTML]{393939} 0.456}          & {\color[HTML]{393939} \textbf{0.204}}  & {\color[HTML]{393939} \textbf{0.150}}  \\ \hline
\textbf{ECSS (Proposed)}                & {\color[HTML]{393939} \textbf{3.506 $\pm$ 0.022}} & {\color[HTML]{393939} \textbf{3.619 $\pm$ 0.028}} & {\color[HTML]{393939} \textbf{0.654}} & {\color[HTML]{393939} \textbf{0.455}} & {\color[HTML]{393939} \textbf{0.215*}} & {\color[HTML]{393939} \textbf{0.152*}} \\ \hline
\qquad{\color[HTML]{393939} w/o emotion}        &   3.424 $\pm$ 0.018   & 3.496 $\pm$ 0.026       & {\color[HTML]{393939} 0.658}          & {\color[HTML]{393939} 0.467}           & {\color[HTML]{393939} 0.224}          & {\color[HTML]{393939} \textbf{0.150}}  \\
\qquad{\color[HTML]{393939} w/o intensity}     &   3.487 $\pm$ 0.035   & 3.523 $\pm$ 0.024  & {\color[HTML]{393939} 0.660}          & {\color[HTML]{393939} \textbf{0.453}}  & {\color[HTML]{393939} 0.210}          & {\color[HTML]{393939} 0.157}           \\
\qquad{\color[HTML]{393939} w/o speaker}      &    3.401 $\pm$ 0.019  &   3.511 $\pm$ 0.027       & {\color[HTML]{393939} 0.666}          & {\color[HTML]{393939} 0.456}           & {\color[HTML]{393939} 0.231}          & {\color[HTML]{393939} 0.152}           \\
\qquad{\color[HTML]{393939} w/o audio}       &   3.391 $\pm$ 0.023   &   3.505 $\pm$ 0.022      & {\color[HTML]{393939} 0.656}          & {\color[HTML]{393939} 0.457}           & {\color[HTML]{393939} \textbf{0.198}} & {\color[HTML]{393939} 0.154}           \\  
\qquad{\color[HTML]{393939} w/o $\mathcal{L}^{cl}$}    &  3.388 $\pm$ 0.025    &   3.312 $\pm$ 0.031             & {\color[HTML]{393939} 0.665}          & {\color[HTML]{393939} 0.459}          & {\color[HTML]{393939} 0.222}          & {\color[HTML]{393939} 0.156}          \\
\hline
\end{tabular}}
\vspace{-2mm}
\end{table*}

\section{Experiments and Results}
\subsubsection{Dataset}
We validate the ECSS on a recently public dataset for conversational speech synthesis called DailyTalk \cite{lee2023dailytalk}, 
that consists of 23,773 audio clips representing 20 hours in total, in which 2,541 conversations were sampled, modified, and recorded. All dialogues are long enough to represent the context of each conversation. The dataset was recorded by a male and a female simultaneously. All speech samples are sampled at 44.10 kHz and coded in 16 bits. We partition the data into training, validation, and test sets at a ratio of 8:1:1.

To obtain the multi-source knowledge,
we invited a professional practitioner to perform fine-grained labeling of DailyTalk data for emotion category and intensity labels while listening to speech and understanding the semantics of utterance. The final distribution of the data is as follows, the number of each of the 7 emotion category labels (happy, sad, angry, disgust, fear, surprise, neutral) are 3871, 722, 226, 186, 74, 497 and 18,197, and the number of each of the 3 emotion intensity labels (weak, medium, strong) are 19,973, 3,646 and 154.


\vspace{-1.5mm}
\subsubsection{Experimental Setup}
In the heterogeneous graph-based emotion context encoder, the dimension of the text node representation $f_{u_j}$ is set to 512, and the dimensions of the remaining type node representations $f_{e_j}$,$f_{i_j}$,$f_{s_j}$, and $f_{a_j}$ are all set to 256. For multi-head attention-based methods, we set the head number as 8.
For the emotion predictor of emotion renderer, the convolutional layer has a convolutional kernel of 3, the LSTM input dimension is 384, the hidden state size is 256, and the forward and backward outputs from the last time steps of the LSTM are spliced using concat before going into the linear layer. For the intensity predictor of emotion renderer, the mean pooling layer convolution kernel size is 2. For the prosody predictor of emotion renderer, the dimensions of the $Query$, $Key$ and $Value$ and output feature $\mathcal{H}^{p}_\mathcal{C}$ are 512, 384, 384 and 256.
The acoustic decoder is configured with reference to FastSpeech2 \cite{ren2021fastspeech}. we use Adam optimizer with $\beta_1$ = 0.9, $\beta_2$ = 0.98. 
For text input, we adopt Grapheme-to-Phoneme (G2P) toolkit\footnote{https://www.github.com/kyubyong/g2p} to convert all text into its phoneme sequence. All speech samples are re-sampled to 22.05 kHz. The mel-spectrum features are extracted with a window length of 25ms and a shift of 10ms. The model is trained on a Tesla V100 GPU with a batch size of 16 and 600k steps.
During ECSS training, the context or dialogue history length is set to 10.
More detailed experimental settings are accessed in the \textit{Appendix} section.

\vspace{-1.5mm}
\subsubsection{Evaluation Metrics}
For the subjective evaluation metrics, we organized a dialogue-level Mean Opinion Score (DMOS) \cite{streijl2016mean} listening test with 30 master students whose second language is English and provided specialized training on the rules to all listeners. Given the dialogue history, they were asked to rate the naturalness DMOS (N-DMOS) and emotional DMOS (E-DMOS) of the synthesized speech of the current utterance, ranging from 1 to 5. Each listener was access to 50 audio samples.
Note that the N-DMOS focuses on the speaking prosody, while the E-DMOS focuses on the richness of the emotional expression of the current utterance and whether it matches the emotional expression of the context.

For the objective evaluation metrics, we calculate the Mean Absolute Error (MAE) between the predicted and real acoustic features to assess the emotional expressiveness of the synthesized audio. Specifically, we assess the acoustic feature in terms of mel-spectrum, pitch, energy, and duration with MAE-M, MAE-P, MAE-E and MAE-D.
In addition, we conduct a visualization study with the help of a third-party Speech Emotion Recognition (SER) model, that was used to identify the emotion categories of the synthesized emotional conversational speech. We plot the confusion matrix to validate the ECSS.

\vspace{-1.6mm}
\subsubsection{Comparative Study}
To demonstrate the effectiveness of our ECSS, we employ three advanced approaches which also employ FastSpeech2 as the TTS backbone as the baseline systems. 

\begin{itemize}
\vspace{-1mm}
    \item \textbf{No emotional context modeling}. The first baseline approach is a vanilla FastSpeech2 \cite{ren2021fastspeech} with no context modeling, which is also representative of state-of-the-art non-conversational TTS systems.
\vspace{-1mm}
    \item \textbf{GRU-based context modeling}. This method involves only text modality and uses an RNN-based uni-directional GRU network to model contextual dependencies in a dialogue sequentially \cite{DBLP:journals/corr/abs-2005-10438}.
    \vspace{-1mm}
    \item  \textbf{Homogeneous Graph-based emotion context modeling}. In the homogeneous graph-based approach \cite{li2022inferring}, each past utterance in the conversation is represented as a node in the graph. Each node is initialized with the corresponding multi-modal features. To achieve emotion rendering using this method, we can extract the graph-enhanced feature of all nodes in dialogue history to predict the emotion and intensity information.
\end{itemize}
\vspace{-2mm}
\subsubsection{Main Results}

As shown in the first five rows of Table \ref{tab:methods}, the ECSS achieves state-of-the-art performance on average, which obtains the optimum results in MAE-M (0.654) and MAE-P (0.455), and suboptimal results in MAE-E (0.215) and MAE-D (0.152).
However, objective experiments may not fully reflect human feelings. By observing the subjective results,
the proposed ECSS model outperforms all baselines with an N-DMOS score of 3.506 and an E-DMOS score of 3.619, which reflects the superiority of our ECSS. ECSS contributes to adequate emotion understanding based on heterogeneous graph context modeling, in which the emotion renderer fully mines the emotion cues to infer the emotion expression state of the current sentence, thus achieving satisfactory emotion rendering effects for conversational speech synthesis.

\subsubsection{Ablation Results}
To evaluate the individual effects of various heterogeneous nodes, including emotion, intensity, speaker and audio, in ECG and the contrastive learning loss $\mathcal{L}^{cl}$ for emotion renderer, we remove these components to build various systems and conduct a series of ablation experiments, and the subjective and objective results are shown in the rows 6 through 10 of Table \ref{tab:methods}.

We can find that removing different types of nodes in the heterogeneous ECG brought about a decrease in objective metrics performance in the vast majority of metrics, and the subjective DMOS scores also showed a drop. For example, after removing the emotion node, the MAE-M, MAE-P and MAE-E decreased by 0.004, 0.012 and 0.009 respectively, while N-DMOS and E-DMOS decreased by 0.082 and 0.123. This suggests that our heterogeneous graph nodes can learn the complex emotional dependencies in the dialog history and achieve full emotional understanding. 
In addition, to validate the $\mathcal{L}^{cl}$, we replace it with the cross-entropy loss. As shown in the last row of Table \ref{tab:methods}, all subjective and objective values are reduced after removing the $\mathcal{L}^{cl}$. This demonstrates that the contrastive learning strategy allows the emotion renderer to distinguish between different emotion categories and intensities better.

\begin{table}[t!]
\caption{\label{tab:length}\textcolor{black}{Objective results of various context lengths.} }
\vspace{-3mm}
\resizebox{0.96\linewidth}{!}{
\begin{tabular}{ccccc}
\hline
{\color[HTML]{393939} \textbf{Length}} & {\color[HTML]{393939} \textbf{MAE-M} ($\downarrow$)} & {\color[HTML]{393939} \textbf{MAE-P} ($\downarrow$)}  & {\color[HTML]{393939} \textbf{MAE-E} ($\downarrow$)} & {\color[HTML]{393939} \textbf{MAE-D} ($\downarrow$)} \\ \hline
{\color[HTML]{393939} 2}      & {\color[HTML]{393939} 0.667}          & {\color[HTML]{393939} 0.468}           & {\color[HTML]{393939} 0.233}          & {\color[HTML]{393939} 0.160}          \\
{\color[HTML]{393939} 3}               & {\color[HTML]{393939} 0.664}          & {\color[HTML]{393939} 0.461}           & {\color[HTML]{393939} 0.229}          & {\color[HTML]{393939} 0.157}          \\
{\color[HTML]{393939} 6}               & {\color[HTML]{393939} 0.662}          & {\color[HTML]{393939} 0.456}           & {\color[HTML]{393939} 0.223}          & {\color[HTML]{393939} 0.155}          \\
9                                      & 0.658                                 & \textbf{0.453}                         & 0.219                                 & 0.153                                 \\
{\color[HTML]{393939} \textbf{10}}     & {\color[HTML]{393939} \textbf{0.654}} & {\color[HTML]{393939} \textbf{0.455*}} & {\color[HTML]{393939} \textbf{0.215}} & {\color[HTML]{393939} \textbf{0.152}} \\
13                                     & 0.660                                 & 0.456                                  & 0.220                                 & 0.154                                 \\
14                                     & 0.661                                 & 0.457                                  & 0.221                                 & 0.153                                 \\ \hline
\end{tabular}}
\vspace{-2.5mm}
\end{table}

\begin{figure}[t!]
\centering
\centerline{\includegraphics[width=\linewidth]{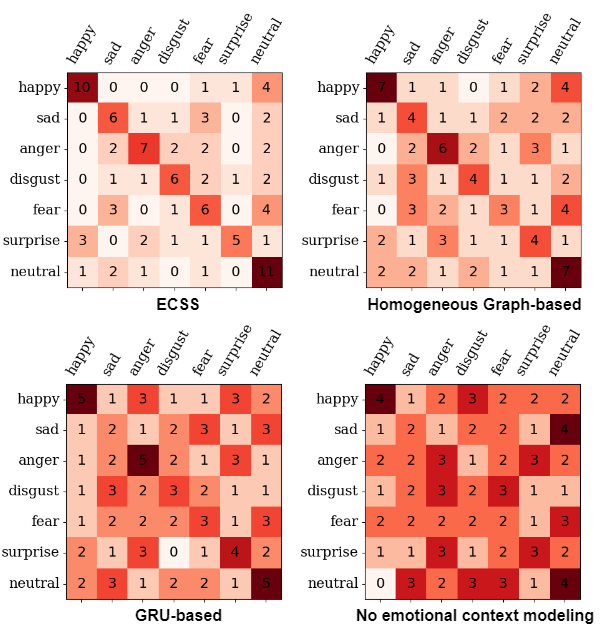}}
\vspace{-2mm}
\caption{The confusion matrix results about the emotion category rendering with the help of speech emotion recognition. The X-axis and Y-axis of subfigures represent perceived and true emotion categories.}
\vspace{-3mm}
\label{fig3}
\end{figure}

\vspace{-2mm}
\subsubsection{Context Length Analysis}
We also explore the effectiveness of emotional context modeling with different context lengths.
Specifically, considering the average number 9.3 of dialogue turns in the DailyTalk, we set the utterance length of dialogue history ranging from 2 to 14 to compare the objective performance. As shown in Table \ref{tab:length}, from a general view, all values decrease when enlarging the context length from 2 to 10, and increase from 10 to 14. This shows that either insufficient or redundant context information will interfere the understanding of emotion cues in context.
 
\subsubsection{Visualization Study}

\begin{figure}[t!]
\centering
\centerline{\includegraphics[width=0.75\linewidth]{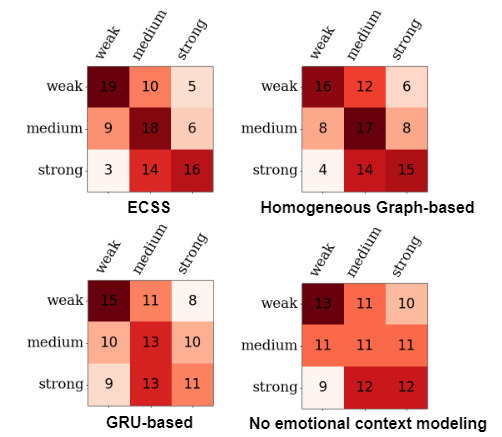}}
\vspace{-3mm}
\caption{The confusion matrix results about the emotion {intensity} rendering. The X-axis and Y-axis of subfigures represent perceived and true {intensity} categories.}
\vspace{-3.5mm}
\label{fig4}
\end{figure}

To demonstrate the emotional expressiveness of synthesized speech more visually, we employ a pre-trained SER model\footnote{https://huggingface.co/ehcalabres/wav2vec2-lg-xlsr-en-speech-emotion-recognition} to identify the emotion category of 400 audio samples synthesized from the ECSS and all baselines, respectively. As shown in Fig. \ref{fig3}, we plot the confusion matrices to show the gap between the different systems. It can be seen that our ECSS outperforms all baselines by presenting a clear diagonal line in the confusion matrix. It proves that the emotional conversational speech synthesized by ECSS performs a clear emotional expression.
In addition, we further conduct a visualization study to validate the emotion intensity rendering. Specifically, five listeners were invited and asked to rate the emotion intensity labels of 400 audios. As shown in Fig. \ref{fig4}, the confusion matrix suggests that the emotional conversational speech synthesized by ECSS performs a clear emotional intensity expression.
The above results also provide further evidence that our ECSS draws on the contextual modeling capabilities of heterogeneous graph and the effective constraints of contrastive learning, thus leading to remarkable performance in both emotion and emotion intensity rendering.

\vspace{-1mm}
\section{Conclusion}
To improve the emotion understanding and rendering in CSS systems, we present a novel ECSS model whereby a heterogeneous Emotional Conversational Graph (ECG) armed with multi-source knowledge in context is used for emotional context modeling, and the emotion renderer with contrastive learning constraint to achieve accurate emotional style inference. Experimental results demonstrate the superiority of ECSS over state-of-the-art CSS systems. The contribution of heterogeneous nodes in ECG and emotion renderer are further demonstrated in ablation studies. To the best of our knowledge, ECSS is the first in-depth conversational speech synthesis study that models emotional expressiveness.  We hope that our work will serve as a basis for future emotional CSS studies.

\section{Acknowledgments}
The research by Rui Liu was funded by 
the Young Scientists Fund of the National Natural Science Foundation of China (No. 62206136), and Guangdong Provincial Key Laboratory of Human Digital Twin (No. 2022B1212010004).
The research by Yifan Hu was funded by the Research and Innovation Projects for Graduate Students in Inner Mongolia Autonomous Region (No. 11200-121028).
The work by Haizhou Li was supported by the National Natural Science Foundation of China (Grant No. 62271432), and Shenzhen Science and Technology Research Fund (Fundamental Research Key Project Grant No. JCYJ20220818103001002).

\bigskip

\bibliography{aaai24}

\begin{thebibliography}{38}
\providecommand{\natexlab}[1]{#1}

\bibitem[{Busso et~al.(2008)Busso, Bulut, Lee, Kazemzadeh, Mower, Kim, Chang,
  Lee, and Narayanan}]{busso2008iemocap}
Busso, C.; Bulut, M.; Lee, C.-C.; Kazemzadeh, A.; Mower, E.; Kim, S.; Chang,
  J.~N.; Lee, S.; and Narayanan, S.~S. 2008.
\newblock IEMOCAP: Interactive emotional dyadic motion capture database.
\newblock \emph{Language resources and evaluation}, 42: 335--359.

\bibitem[{Dias et~al.(2022)Dias, Andalo, Padilha, Bertocco, Almeida, Costa, and
  Rocha}]{dias2022cross}
Dias, W.; Andalo, F.; Padilha, R.; Bertocco, G.; Almeida, W.; Costa, P.; and
  Rocha, A. 2022.
\newblock Cross-dataset emotion recognition from facial expressions through
  convolutional neural networks.
\newblock \emph{Journal of Visual Communication and Image Representation}, 82:
  103395.

\bibitem[{Firdaus et~al.(2020)Firdaus, Chauhan, Ekbal, and
  Bhattacharyya}]{firdaus-etal-2020-meisd}
Firdaus, M.; Chauhan, H.; Ekbal, A.; and Bhattacharyya, P. 2020.
\newblock {MEISD}: A Multimodal Multi-Label Emotion, Intensity and Sentiment
  Dialogue Dataset for Emotion Recognition and Sentiment Analysis in
  Conversations.
\newblock In \emph{Proceedings of the 28th International Conference on
  Computational Linguistics}, 4441--4453. Barcelona, Spain (Online):
  International Committee on Computational Linguistics.

\bibitem[{Ghosal et~al.(2019)Ghosal, Majumder, Poria, Chhaya, and
  Gelbukh}]{ghosal-etal-2019-dialoguegcn}
Ghosal, D.; Majumder, N.; Poria, S.; Chhaya, N.; and Gelbukh, A. 2019.
\newblock {D}ialogue{GCN}: A Graph Convolutional Neural Network for Emotion
  Recognition in Conversation.
\newblock In \emph{Proceedings of the 2019 Conference on Empirical Methods in
  Natural Language Processing and the 9th International Joint Conference on
  Natural Language Processing (EMNLP-IJCNLP)}, 154--164. Hong Kong, China:
  Association for Computational Linguistics.

\bibitem[{Goel et~al.(2021)Goel, Susan, Vashisht, and Dhanda}]{goel2021emotion}
Goel, R.; Susan, S.; Vashisht, S.; and Dhanda, A. 2021.
\newblock Emotion-aware transformer encoder for empathetic dialogue generation.
\newblock In \emph{2021 9th International Conference on Affective Computing and
  Intelligent Interaction Workshops and Demos (ACIIW)}, 1--6. IEEE.

\bibitem[{Guo et~al.(2020)Guo, Zhang, Soong, He, and
  Xie}]{DBLP:journals/corr/abs-2005-10438}
Guo, H.; Zhang, S.; Soong, F.~K.; He, L.; and Xie, L. 2020.
\newblock Conversational End-to-End {TTS} for Voice Agent.
\newblock \emph{CoRR}, abs/2005.10438.

\bibitem[{Hu et~al.(2020)Hu, Dong, Wang, and Sun}]{hu2020heterogeneous}
Hu, Z.; Dong, Y.; Wang, K.; and Sun, Y. 2020.
\newblock Heterogeneous graph transformer.
\newblock In \emph{Proceedings of the web conference 2020}, 2704--2710.

\bibitem[{Huang et~al.(2022)Huang, Ren, Liu, Cui, and
  Zhao}]{huang2022generspeech}
Huang, R.; Ren, Y.; Liu, J.; Cui, C.; and Zhao, Z. 2022.
\newblock Generspeech: Towards style transfer for generalizable out-of-domain
  text-to-speech.
\newblock \emph{Advances in Neural Information Processing Systems}, 35:
  10970--10983.

\bibitem[{Im et~al.(2022)Im, Lee, Kim, and Lee}]{9747098}
Im, C.-B.; Lee, S.-H.; Kim, S.-B.; and Lee, S.-W. 2022.
\newblock EMOQ-TTS: Emotion Intensity Quantization for Fine-Grained
  Controllable Emotional Text-to-Speech.
\newblock In \emph{ICASSP 2022 - 2022 IEEE International Conference on
  Acoustics, Speech and Signal Processing (ICASSP)}, 6317--6321.

\bibitem[{Khosla et~al.(2020)Khosla, Teterwak, Wang, Sarna, Tian, Isola,
  Maschinot, Liu, and Krishnan}]{DBLP:journals/corr/abs-2004-11362}
Khosla, P.; Teterwak, P.; Wang, C.; Sarna, A.; Tian, Y.; Isola, P.; Maschinot,
  A.; Liu, C.; and Krishnan, D. 2020.
\newblock Supervised Contrastive Learning.
\newblock \emph{CoRR}, abs/2004.11362.

\bibitem[{Kim, Kong, and Son(2021)}]{kim2021conditional}
Kim, J.; Kong, J.; and Son, J. 2021.
\newblock Conditional variational autoencoder with adversarial learning for
  end-to-end text-to-speech.
\newblock In \emph{International Conference on Machine Learning}, 5530--5540.
  PMLR.

\bibitem[{Kong, Kim, and Bae(2020)}]{kong2020hifi}
Kong, J.; Kim, J.; and Bae, J. 2020.
\newblock Hifi-gan: Generative adversarial networks for efficient and high
  fidelity speech synthesis.
\newblock \emph{Advances in Neural Information Processing Systems}, 33:
  17022--17033.

\bibitem[{Lee, Park, and Kim(2023)}]{lee2023dailytalk}
Lee, K.; Park, K.; and Kim, D. 2023.
\newblock Dailytalk: Spoken dialogue dataset for conversational text-to-speech.
\newblock In \emph{ICASSP 2023-2023 IEEE International Conference on Acoustics,
  Speech and Signal Processing (ICASSP)}, 1--5. IEEE.

\bibitem[{Li et~al.(2022{\natexlab{a}})Li, Meng, Li, Wu, Meng, Weng, and
  Su}]{li2022enhancing}
Li, J.; Meng, Y.; Li, C.; Wu, Z.; Meng, H.; Weng, C.; and Su, D.
  2022{\natexlab{a}}.
\newblock Enhancing speaking styles in conversational text-to-speech synthesis
  with graph-based multi-modal context modeling.
\newblock In \emph{ICASSP 2022-2022 IEEE International Conference on Acoustics,
  Speech and Signal Processing (ICASSP)}, 7917--7921. IEEE.

\bibitem[{Li et~al.(2022{\natexlab{b}})Li, Meng, Wu, Wu, Jia, Meng, Tian, Wang,
  and Wang}]{li2022inferring}
Li, J.; Meng, Y.; Wu, X.; Wu, Z.; Jia, J.; Meng, H.; Tian, Q.; Wang, Y.; and
  Wang, Y. 2022{\natexlab{b}}.
\newblock Inferring speaking styles from multi-modal conversational context by
  multi-scale relational graph convolutional networks.
\newblock In \emph{Proceedings of the 30th ACM International Conference on
  Multimedia}, 5811--5820.

\bibitem[{Li et~al.(2019)Li, Liu, Liu, Zhao, and Liu}]{li2019neural}
Li, N.; Liu, S.; Liu, Y.; Zhao, S.; and Liu, M. 2019.
\newblock Neural speech synthesis with transformer network.
\newblock In \emph{Proceedings of the AAAI conference on artificial
  intelligence}, volume~33, 6706--6713.

\bibitem[{Li et~al.(2022{\natexlab{c}})Li, Wang, Xie, Wang, and Xie}]{9747987}
Li, T.; Wang, X.; Xie, Q.; Wang, Z.; and Xie, L. 2022{\natexlab{c}}.
\newblock Cross-Speaker Emotion Disentangling and Transfer for End-to-End
  Speech Synthesis.
\newblock \emph{IEEE/ACM Transactions on Audio, Speech, and Language
  Processing}, 30: 1448--1460.

\bibitem[{Li et~al.(2022{\natexlab{d}})Li, Tu, Liang, and
  Xu}]{li2022developing}
Li, Z.; Tu, G.; Liang, X.; and Xu, R. 2022{\natexlab{d}}.
\newblock Developing Relationships: A Heterogeneous Graph Network with
  Learnable Edge Representation for Emotion Identification in Conversations.
\newblock In \emph{CAAI International Conference on Artificial Intelligence},
  310--322. Springer.

\bibitem[{Liu et~al.(2021{\natexlab{a}})Liu, Sisman, Bao, Yang, Gao, and
  Li}]{9271923}
Liu, R.; Sisman, B.; Bao, F.; Yang, J.; Gao, G.; and Li, H. 2021{\natexlab{a}}.
\newblock Exploiting Morphological and Phonological Features to Improve
  Prosodic Phrasing for Mongolian Speech Synthesis.
\newblock \emph{IEEE/ACM Transactions on Audio, Speech, and Language
  Processing}, 29: 274--285.

\bibitem[{Liu et~al.(2021{\natexlab{b}})Liu, Sisman, Gao, and Li}]{9420276}
Liu, R.; Sisman, B.; Gao, G.; and Li, H. 2021{\natexlab{b}}.
\newblock Expressive TTS Training With Frame and Style Reconstruction Loss.
\newblock \emph{IEEE/ACM Transactions on Audio, Speech, and Language
  Processing}, 29: 1806--1818.

\bibitem[{Liu et~al.(2022{\natexlab{a}})Liu, Sisman, Gao, and Li}]{9767637}
Liu, R.; Sisman, B.; Gao, G.; and Li, H. 2022{\natexlab{a}}.
\newblock Decoding Knowledge Transfer for Neural Text-to-Speech Training.
\newblock \emph{IEEE/ACM Transactions on Audio, Speech, and Language
  Processing}, 30: 1789--1802.

\bibitem[{Liu et~al.(2021{\natexlab{c}})Liu, Sisman, Lin, and
  Li}]{liu2021fasttalker}
Liu, R.; Sisman, B.; Lin, Y.; and Li, H. 2021{\natexlab{c}}.
\newblock Fasttalker: A neural text-to-speech architecture with shallow and
  group autoregression.
\newblock \emph{Neural Networks}, 141: 306--314.

\bibitem[{Liu et~al.(2022{\natexlab{b}})Liu, Sisman, Schuller, Gao, and
  Li}]{liu2022accurate}
Liu, R.; Sisman, B.; Schuller, B.; Gao, G.; and Li, H. 2022{\natexlab{b}}.
\newblock Accurate emotion strength assessment for seen and unseen speech based
  on data-driven deep learning.
\newblock \emph{arXiv preprint arXiv:2206.07229}.

\bibitem[{McTear(2022)}]{mctear2022conversational}
McTear, M. 2022.
\newblock \emph{Conversational ai: Dialogue systems, conversational agents, and
  chatbots}.
\newblock Springer Nature.

\bibitem[{Poria et~al.(2018)Poria, Hazarika, Majumder, Naik, Cambria, and
  Mihalcea}]{DBLP:journals/corr/abs-1810-02508}
Poria, S.; Hazarika, D.; Majumder, N.; Naik, G.; Cambria, E.; and Mihalcea, R.
  2018.
\newblock MELD: A Multimodal Multi-Party Dataset for Emotion Recognition in
  Conversations.
\newblock \emph{CoRR}, abs/1810.02508.

\bibitem[{Ren et~al.(2021)Ren, Hu, Tan, Qin, Zhao, Zhao, and
  Liu}]{ren2021fastspeech}
Ren, Y.; Hu, C.; Tan, X.; Qin, T.; Zhao, S.; Zhao, Z.; and Liu, T.-Y. 2021.
\newblock FastSpeech 2: Fast and High-Quality End-to-End Text to Speech.
\newblock In \emph{International Conference on Learning Representations}.

\bibitem[{Saganowski et~al.(2022)Saganowski, Komoszy{\'n}ska, Behnke, Perz,
  Kunc, Klich, Kaczmarek, and Kazienko}]{saganowski2022emognition}
Saganowski, S.; Komoszy{\'n}ska, J.; Behnke, M.; Perz, B.; Kunc, D.; Klich, B.;
  Kaczmarek, {\L}.~D.; and Kazienko, P. 2022.
\newblock Emognition dataset: emotion recognition with self-reports, facial
  expressions, and physiology using wearables.
\newblock \emph{Scientific data}, 9(1): 158.

\bibitem[{Seaborn et~al.(2021)Seaborn, Miyake, Pennefather, and
  Otake-Matsuura}]{seaborn2021voice}
Seaborn, K.; Miyake, N.~P.; Pennefather, P.; and Otake-Matsuura, M. 2021.
\newblock Voice in human--agent interaction: A survey.
\newblock \emph{ACM Computing Surveys (CSUR)}, 54(4): 1--43.

\bibitem[{Song et~al.(2023)Song, Giunchiglia, Shi, Shen, and
  Xu}]{song2023sunet}
Song, R.; Giunchiglia, F.; Shi, L.; Shen, Q.; and Xu, H. 2023.
\newblock SUNET: Speaker-utterance interaction Graph Neural Network for Emotion
  Recognition in Conversations.
\newblock \emph{Engineering Applications of Artificial Intelligence}, 123:
  106315.

\bibitem[{Streijl, Winkler, and Hands(2016)}]{streijl2016mean}
Streijl, R.~C.; Winkler, S.; and Hands, D.~S. 2016.
\newblock Mean opinion score (MOS) revisited: methods and applications,
  limitations and alternatives.
\newblock \emph{Multimedia Systems}, 22(2): 213--227.

\bibitem[{Tulshan and Dhage(2019)}]{tulshan2019survey}
Tulshan, A.~S.; and Dhage, S.~N. 2019.
\newblock Survey on virtual assistant: Google assistant, siri, cortana, alexa.
\newblock In \emph{Advances in Signal Processing and Intelligent Recognition
  Systems: 4th International Symposium SIRS 2018, Bangalore, India, September
  19--22, 2018, Revised Selected Papers 4}, 190--201. Springer.

\bibitem[{Wang et~al.(2017)Wang, Skerry-Ryan, Stanton, Wu, Weiss, Jaitly, Yang,
  Xiao, Chen, Bengio et~al.}]{wang2017tacotron}
Wang, Y.; Skerry-Ryan, R.; Stanton, D.; Wu, Y.; Weiss, R.~J.; Jaitly, N.; Yang,
  Z.; Xiao, Y.; Chen, Z.; Bengio, S.; et~al. 2017.
\newblock Tacotron: Towards End-to-End Speech Synthesis.
\newblock \emph{Proc. Interspeech 2017}, 4006--4010.

\bibitem[{Wang et~al.(2018)Wang, Stanton, Zhang, Ryan, Battenberg, Shor, Xiao,
  Jia, Ren, and Saurous}]{wang2018style}
Wang, Y.; Stanton, D.; Zhang, Y.; Ryan, R.-S.; Battenberg, E.; Shor, J.; Xiao,
  Y.; Jia, Y.; Ren, F.; and Saurous, R.~A. 2018.
\newblock Style tokens: Unsupervised style modeling, control and transfer in
  end-to-end speech synthesis.
\newblock In \emph{International conference on machine learning}, 5180--5189.
  PMLR.

\bibitem[{Xue et~al.(2023)Xue, Deng, Wang, Li, Gao, Tao, Sun, and
  Liang}]{xue2023m}
Xue, J.; Deng, Y.; Wang, F.; Li, Y.; Gao, Y.; Tao, J.; Sun, J.; and Liang, J.
  2023.
\newblock M 2-CTTS: End-to-End Multi-Scale Multi-Modal Conversational
  Text-to-Speech Synthesis.
\newblock In \emph{ICASSP 2023-2023 IEEE International Conference on Acoustics,
  Speech and Signal Processing (ICASSP)}, 1--5. IEEE.

\bibitem[{Zhong, Wang, and Miao(2019)}]{DBLP:journals/corr/abs-1909-10681}
Zhong, P.; Wang, D.; and Miao, C. 2019.
\newblock Knowledge-Enriched Transformer for Emotion Detection in Textual
  Conversations.
\newblock \emph{CoRR}, abs/1909.10681.

\bibitem[{Zhou et~al.(2023)Zhou, Sisman, Rana, Schuller, and Li}]{9778970}
Zhou, K.; Sisman, B.; Rana, R.; Schuller, B.~W.; and Li, H. 2023.
\newblock Emotion Intensity and its Control for Emotional Voice Conversion.
\newblock \emph{IEEE Transactions on Affective Computing}, 14(1): 31--48.

\bibitem[{Zhou et~al.(2020)Zhou, Gao, Li, and Shum}]{zhou2020design}
Zhou, L.; Gao, J.; Li, D.; and Shum, H.-Y. 2020.
\newblock The design and implementation of xiaoice, an empathetic social
  chatbot.
\newblock \emph{Computational Linguistics}, 46(1): 53--93.

\bibitem[{Zuo et~al.(2023)Zuo, Liu, Zhao, Gao, and Li}]{10095836}
Zuo, H.; Liu, R.; Zhao, J.; Gao, G.; and Li, H. 2023.
\newblock Exploiting Modality-Invariant Feature for Robust Multimodal Emotion
  Recognition with Missing Modalities.
\newblock In \emph{ICASSP 2023 - 2023 IEEE International Conference on
  Acoustics, Speech and Signal Processing (ICASSP)}, 1--5.

\end{thebibliography}

\appendix

\section{Appendix. Detailed Experimental Settings}
\label{app:config}


We list the detailed model setup of ECSS. The detailed experimental settings of the \textit{Heterogeneous Graph-based Emotional Context Encoder} and \textit{Emotional Conversational Speech Synthesizer} are shown in Table \ref{tab:setup}.

\begin{table}[th!]
\caption{\label{tab:setup}Detailed Model Setup of ECSS.}
\resizebox{\linewidth}{!}{
\begin{tabular}{c|cc|c}
\hline

\multicolumn{4}{c}{\textbf{ECSS Model Setup}}                                                                                                                                                                                                                                                                           \\ \hline
\multicolumn{1}{c|}{}                                                                                                                    & \multicolumn{1}{c|}{{\color[HTML]{393939} }}                                                                               & \multicolumn{1}{c|}{{\color[HTML]{393939} Text Node Representation}}                                                                                 & {\color[HTML]{393939} 512} \\ \cline{3-4} 
\multicolumn{1}{c|}{}                                                                                                                    & \multicolumn{1}{c|}{{\color[HTML]{393939} }}                                                                               & \multicolumn{1}{c|}{{\color[HTML]{393939} \begin{tabular}[c]{@{}c@{}}Emotion,Intensity,\\ Speaker and \\ Audio Node \\ Representation\end{tabular}}} & {\color[HTML]{393939} 256} \\ \cline{3-4} 
\multicolumn{1}{c|}{}                                                                                                                    & \multicolumn{1}{c|}{{\color[HTML]{393939} }}                                                                               & \multicolumn{1}{c|}{{\color[HTML]{393939} HGTConv Hidden Channels}}                                                                                  & {\color[HTML]{393939} 384} \\ \cline{3-4} 
\multicolumn{1}{c|}{}                                                                                                                    & \multicolumn{1}{c|}{{\color[HTML]{393939} }}                                                                               & \multicolumn{1}{c|}{HGTConv Head Number}                                                                                                             & 2                          \\ \cline{3-4} 
\multicolumn{1}{c|}{\multirow{-8}{*}{\begin{tabular}[c]{@{}c@{}}Heterogeneous\\ Graph-based\\ Emotional\\ Context Encoder\end{tabular}}} & \multicolumn{1}{c|}{\multirow{-8}{*}{{\color[HTML]{393939} Heterogeneous ECG}}}                                            & \multicolumn{1}{c|}{{\color[HTML]{393939} HGTConv Layer}}                                                                                            & {\color[HTML]{393939} 1}   \\ \hline
\multicolumn{1}{c|}{}                                                                                                                    & \multicolumn{1}{c|}{}                                                                                                      & \multicolumn{1}{c|}{ConvBlock2D Kernel}                                                                                                              & 3                          \\ \cline{3-4} 
\multicolumn{1}{c|}{}                                                                                                                    & \multicolumn{1}{c|}{}                                                                                                      & \multicolumn{1}{c|}{ConvBlock2D Layer}                                                                                                               & 2                          \\ \cline{3-4} 
\multicolumn{1}{c|}{}                                                                                                                    & \multicolumn{1}{c|}{}                                                                                                      & \multicolumn{1}{c|}{LSTM Input Dimension}                                                                                                            & 384                        \\ \cline{3-4} 
\multicolumn{1}{c|}{}                                                                                                                    & \multicolumn{1}{c|}{}                                                                                                      & \multicolumn{1}{c|}{LSTM Hidden State Size}                                                                                                          & 256                        \\ \cline{3-4} 
\multicolumn{1}{c|}{}                                                                                                                    & \multicolumn{1}{c|}{\multirow{-5}{*}{\begin{tabular}[c]{@{}c@{}}Emotion Predictor\\ of\\ Emotion Renderer\end{tabular}}}   & \multicolumn{1}{c|}{\begin{tabular}[c]{@{}c@{}}Emotion Predictor\\ Output Dimension\end{tabular}}                                                    & 7                          \\ \cline{2-4} 
\multicolumn{1}{c|}{}                                                                                                                    & \multicolumn{1}{c|}{}                                                                                                      & \multicolumn{1}{c|}{ConvBlock2D Kernel}                                                                                                              & 3                          \\ \cline{3-4} 
\multicolumn{1}{c|}{}                                                                                                                    & \multicolumn{1}{c|}{}                                                                                                      & \multicolumn{1}{c|}{ConvBlock2D Layer}                                                                                                               & 2                          \\ \cline{3-4} 
\multicolumn{1}{c|}{}                                                                                                                    & \multicolumn{1}{c|}{}                                                                                                      & \multicolumn{1}{c|}{LSTM Input Dimension}                                                                                                            & 384                        \\ \cline{3-4} 
\multicolumn{1}{c|}{}                                                                                                                    & \multicolumn{1}{c|}{}                                                                                                      & \multicolumn{1}{c|}{LSTM Hidden State Size}                                                                                                          & 256                        \\ \cline{3-4} 
\multicolumn{1}{c|}{}                                                                                                                    & \multicolumn{1}{c|}{}                                                                                                      & \multicolumn{1}{c|}{AvgPool2d Kernel}                                                                                                                & 2                          \\ \cline{3-4} 
\multicolumn{1}{c|}{}                                                                                                                    & \multicolumn{1}{c|}{\multirow{-6}{*}{\begin{tabular}[c]{@{}c@{}}Intensity Predictor\\ of\\ Emotion Renderer\end{tabular}}} & \multicolumn{1}{c|}{Output Dimension}                                                                                                                & 3                          \\ \cline{2-4} 
\multicolumn{1}{c|}{}                                                                                                                    & \multicolumn{1}{c|}{}                                                                                                      & \multicolumn{1}{c|}{Query Dimension}                                                                                                                 & 512                        \\ \cline{3-4} 
\multicolumn{1}{c|}{}                                                                                                                    & \multicolumn{1}{c|}{}                                                                                                      & \multicolumn{1}{c|}{Key and Value Dimension}                                                                                                         & 384                        \\ \cline{3-4} 
\multicolumn{1}{c|}{}                                                                                                                    & \multicolumn{1}{c|}{}                                                                                                      & \multicolumn{1}{c|}{\begin{tabular}[c]{@{}c@{}}Prosody Predictor\\ Head Number\end{tabular}}                                                         & 2                          \\ \cline{3-4} 
\multicolumn{1}{c|}{}                                                                                                                    & \multicolumn{1}{c|}{\multirow{-4}{*}{\begin{tabular}[c]{@{}c@{}}Prosody Predictor\\ of\\ Emotion Renderer\end{tabular}}}   & \multicolumn{1}{c|}{Output Dimension}                                                                                                                & 256                        \\ \cline{2-4} 
\multicolumn{1}{c|}{}                                                                                                                    & \multicolumn{1}{c|}{}                                                                                                      & \multicolumn{1}{c|}{\begin{tabular}[c]{@{}c@{}}Reference Encoder\\ Mel Channels\end{tabular}}                                                        & 80                         \\ \cline{3-4} 
\multicolumn{1}{c|}{}                                                                                                                    & \multicolumn{1}{c|}{}                                                                                                      & \multicolumn{1}{c|}{\begin{tabular}[c]{@{}c@{}}Reference Encoder\\ GRU Size\end{tabular}}                                                            & 128                        \\ \cline{3-4} 
\multicolumn{1}{c|}{}                                                                                                                    & \multicolumn{1}{c|}{}                                                                                                      & \multicolumn{1}{c|}{\begin{tabular}[c]{@{}c@{}}Style Token\\ Embedding Size\end{tabular}}                                                            & 256                        \\ \cline{3-4} 
\multicolumn{1}{c|}{}                                                                                                                    & \multicolumn{1}{c|}{\multirow{-7}{*}{\begin{tabular}[c]{@{}c@{}}Prosody\\ Extractor\end{tabular}}}                         & \multicolumn{1}{c|}{Style Token Layer GSTs}                                                                                                          & 7                          \\ \cline{2-4} 
\multicolumn{1}{c|}{}                                                                                                                    & \multicolumn{1}{c|}{}                                                                                                      & \multicolumn{1}{c|}{Word Vector Dimension}                                                                                                           & 256                        \\ \cline{3-4} 
\multicolumn{1}{c|}{}                                                                                                                    & \multicolumn{1}{c|}{}                                                                                                      & \multicolumn{1}{c|}{Position Encoding Hidden}                                                                                                        & 256                        \\ \cline{3-4} 
\multicolumn{1}{c|}{}                                                                                                                    & \multicolumn{1}{c|}{}                                                                                                      & \multicolumn{1}{c|}{Transformer FFTBlock Layer}                                                                                                      & 4                          \\ \cline{3-4} 
\multicolumn{1}{c|}{}                                                                                                                    & \multicolumn{1}{c|}{}                                                                                                      & \multicolumn{1}{c|}{Transformer FFTBlock Head}                                                                                                       & 2                          \\ \cline{3-4} 
\multicolumn{1}{c|}{}                                                                                                                    & \multicolumn{1}{c|}{\multirow{-5}{*}{Text Encoder}}                                                                        & \multicolumn{1}{c|}{Transformer FFTBlock Dropout}                                                                                                    & 0.2                        \\ \cline{2-4} 
\multicolumn{1}{c|}{}                                                                                                                    & \multicolumn{1}{c|}{Speaker Encoder}                                                                                       & \multicolumn{1}{c|}{Speaker Embedding Size}                                                                                                          & 256                        \\ \cline{2-4} 
\multicolumn{1}{c|}{}                                                                                                                    & \multicolumn{1}{c|}{Feature Aggregator}                                                                                    & \multicolumn{1}{c|}{\begin{tabular}[c]{@{}c@{}}Feature Aggregator\\ Output Dimension\end{tabular}}                                                   & 256                        \\ \cline{2-4} 
\multicolumn{1}{c|}{}                                                                                                                    & \multicolumn{1}{c|}{}                                                                                                      & \multicolumn{1}{c|}{Word Vector Dimension}                                                                                                           & 256                        \\ \cline{3-4} 
\multicolumn{1}{c|}{}                                                                                                                    & \multicolumn{1}{c|}{}                                                                                                      & \multicolumn{1}{c|}{Position Encoding Hidden}                                                                                                        & 256                        \\ \cline{3-4} 
\multicolumn{1}{c|}{}                                                                                                                    & \multicolumn{1}{c|}{}                                                                                                      & \multicolumn{1}{c|}{Transformer FFTBlock Layer}                                                                                                      & 6                          \\ \cline{3-4} 
\multicolumn{1}{c|}{}                                                                                                                    & \multicolumn{1}{c|}{}                                                                                                      & \multicolumn{1}{c|}{{\color[HTML]{393939} Transformer FFTBlock Head}}                                                                                & {\color[HTML]{393939} 2}   \\ \cline{3-4} 
\multicolumn{1}{c|}{}                                                                                                                    & \multicolumn{1}{c|}{}                                                                                                      & \multicolumn{1}{c|}{{\color[HTML]{393939} Transformer FFTBlock Dropout}}                                                                             & {\color[HTML]{393939} 0.2} \\ \cline{3-4} 
\multicolumn{1}{c|}{}                                                                                                                    & \multicolumn{1}{c|}{}                                                                                                      & \multicolumn{1}{c|}{{\color[HTML]{000000} Postnet Embedding Dimension}}                                                                              & {\color[HTML]{393939} 512} \\ \cline{3-4} 
\multicolumn{1}{c|}{\multirow{-40}{*}{\begin{tabular}[c]{@{}c@{}}Emotional\\ Conversational\\ Speech\\ Synthesizer\end{tabular}}}        & \multicolumn{1}{c|}{\multirow{-7}{*}{Acoustic Decoder}}                                                                    & \multicolumn{1}{c|}{Postnet Kernel}                                                                                                                  & 5                          \\ \hline
\end{tabular}
}
\end{table}

\end{document}